\documentclass[letterpaper]{article} 
\usepackage{aaai25}  
\usepackage{times}  
\usepackage{helvet}  
\usepackage{courier}  
\usepackage[hyphens]{url}  
\usepackage{graphicx} 
\usepackage{natbib}  
\usepackage{caption} 
\usepackage{algorithm}
\usepackage{algorithmic}

\usepackage{newfloat}
\usepackage{listings}
\DeclareCaptionStyle{ruled}{labelfont=normalfont,labelsep=colon,strut=off} 
\floatstyle{ruled}
\newfloat{listing}{tb}{lst}{}
\floatname{listing}{Listing}
\title{Exploring Task-Level Optimal Prompts for Visual In-Context Learning}
\author{
    Yan Zhu\textsuperscript{\rm 1}\equalcontrib, Huan Ma\textsuperscript{\rm 1}\equalcontrib, Changqing Zhang\textsuperscript{\rm 1}\thanks{Corresponding author.}
}
\affiliations{
    \textsuperscript{\rm 1}Tianjin University\\

    College of Intelligence and Computing\\
    Tianjin University, China\\
    zhangchangqing@tju.edu.cn
}

\usepackage{bibentry}

\usepackage{epsfig}
\usepackage{graphicx}
\usepackage{subcaption}
\usepackage{amsmath} 
\usepackage{booktabs}
\usepackage{multirow}
\usepackage{xcolor}
\usepackage{xcolor}
\usepackage{bm}
\usepackage{tabularx}
\usepackage{svg}
\newcommand{\best}[1]{\bm{\textcolor{olive}{#1}^1}}
\newcommand{\second}[1]{\bm{\textcolor{blue}{#1}^2}}
\newcommand{\bestcls}[1]{\bm{#1}}

\begin{document}

\maketitle

\begin{abstract}
With the development of Vision Foundation Models (VFMs) in recent years, Visual In-Context Learning (VICL) has become a better choice compared to modifying models in most scenarios. Different from retraining or fine-tuning model, VICL does not require modifications to the model's weights or architecture, and only needs a prompt with demonstrations to teach VFM how to solve tasks. Currently, significant computational cost for finding optimal prompts for every test sample hinders the deployment of VICL, as determining which demonstrations to use for constructing prompts is very costly. In this paper, however, we find a counterintuitive phenomenon that most test samples actually achieve optimal performance under the same prompts, and searching for sample-level prompts only costs more time but results in completely identical prompts. Therefore, we propose task-level prompting to reduce the cost of searching for prompts during the inference stage and introduce two time-saving yet effective task-level prompt search strategies. Extensive experimental results show that our proposed method can identify near-optimal prompts and reach the best VICL performance with a minimal cost that prior work has never achieved.
\end{abstract}

\section{Introduction}
With the development of Vision Foundation Models(VFMs), many tasks in visual scenes no longer require training new models but can be solved by VLFs provided by model service providers, which are more affordable and convenient~\cite{li2023unmasked,wu2023revisiting}. However, the performance of directly deploying VFMs for specific tasks is often unsatisfactory, and task-specific adaptation is necessary for better performance. In comparison to modifying model weights, Visual In-Context Learning (VICL)~\cite{yang2024imagebrush,chen2024visual} is a better choice. Unlike retraining or fine-tuning models, VICL does not necessitate modifying the model's weights or architecture, but only needs to teach VFMs how to solve problems using prompts with a set of demonstrations. However, VICL does not always achieve good results under arbitrary prompts, so determining the appropriate demonstrations to construct prompts is the key challenge in enhancing VICL performance~\cite{li2024visual,dedhia2024promptu}.

\begin{figure*}[ht]
         \centering
         \includegraphics[width=\textwidth]{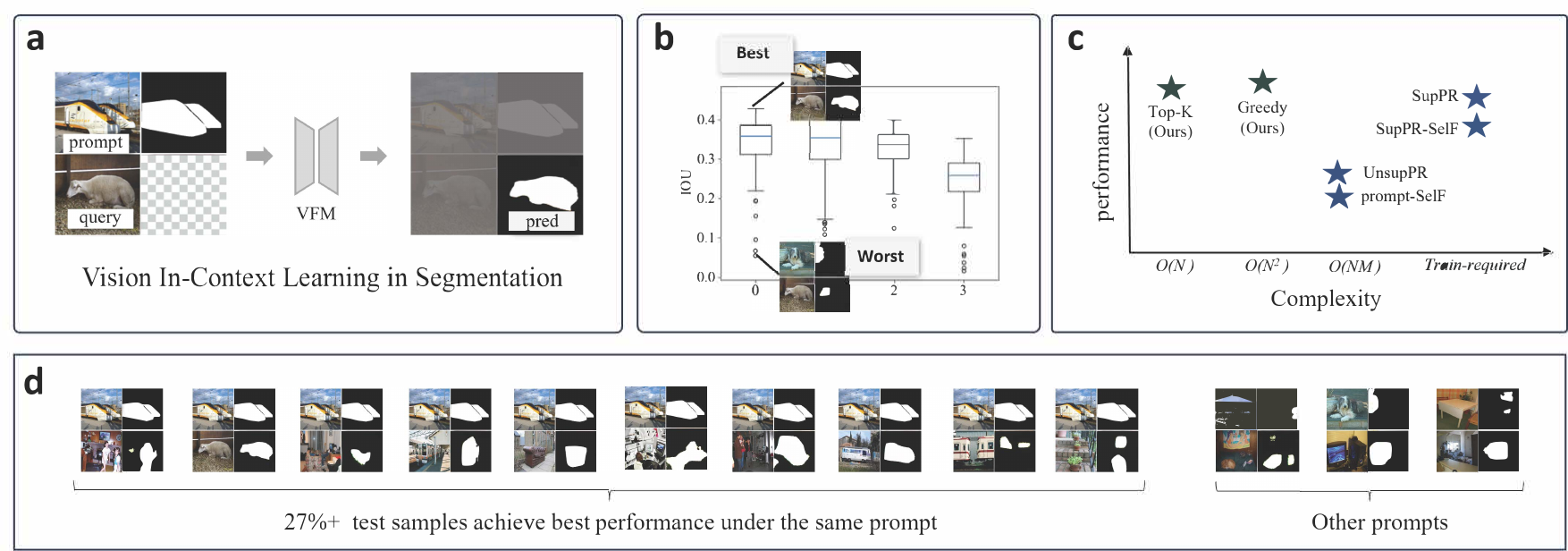}
     \caption{
     (a) The deployment example of VICL in segmentation task. The one-shot case is presented in the figure, and the few-shot prediction is the average prediction under multiple different demonstrations. (b)
     Selecting different prompts makes a significant impact on the tasks. 
     The variance across different prompts is large, even resulting in cases where the metric approaches zero. (c) Comparison of time complexity and performance. Our methods (task-level prompting) significantly reduce complexity while ensuring that the performance is not worse than that of more complex methods. (d) Motivation for task-level prompting. We find that during the testing phase, most samples achieve optimal performance under the same prompt. As shown in the figure, more than 27\% of the samples achieve their best performance under the same prompt, which means that finding the optimal task-level prompt ensures that at least 27\% of the samples obtain the best prompt. In contrast, the sample-level prompt searching strategy only finds the optimal prompt for 15.03\% of the samples (for details please refer to results section).}
     \label{fig:motivation}
 \end{figure*}

Currently, there are primarily two ways to select demonstrations for constructing prompts: the rule-guided strategy and the reward-model-based strategy. The former strategy involves selecting demonstrations based on rules, like UnsupPR~\cite{NEURIPS2023_398ae57e} and prompt-SelF~\cite{sun2023exploring}, which calculate the similarity between the query samples and the training samples using features extracted by the pretrained CLIP~\cite{radford2021learning} model to select prompts. Although this type of method is very simple, these heuristic methods struggle to guarantee performance.
To solve the issue, another strategy is to train a reward model that can predict performance under different prompts. For example, both SupPR~\cite{NEURIPS2023_398ae57e} and InMeMo~\cite{zhang2024instruct} train an additional reward model on the validation set to predict the compatibility between prompts and query samples. However, to obtain such a reward model, it requires a large amount of labeled data and incurs a high computational cost, or the reward model will suffer from severe overfitting. Unfortunately, this high computational cost severely hinders the deployment of VICL.

Is it really necessary to spend a significant amount of computation to find the optimal prompt for each individual sample? Perhaps it is not actually needed. As shown in Fig.~\ref{fig:motivation}(d), 
\emph{we find that under a specific task, the prompts resulting in the best performance for different samples are always the same}. 
The prompt that performs best on the task, even if it cannot achieve optimal performance for some samples, can still obtain relatively good results compared with most prompts. In other words, there is no need to spend a massive amount of computational effort at the risk of overfitting to select different demonstrations to form a prompt for distinct samples (experimental results show that using a single prompt for all samples performs even better than those costly sample-level methods).

In this paper, we propose a reward-based, training-free approach to find the optimal task-level prompt. This strategy not only lessens the time taken during inference but also sustains performance when compared to more time-consuming methods. Specifically, we present two new strategies for effectively searching for the best demonstrations: (1) Top-$K$ strategy and (2) Greedy search strategy. We concentrate on a general setting where a labeled set of size $N$ is provided. The aim of our strategies is to carry out combinatorial optimization over this set to discover optimal demonstrations. In particular, Top-$K$ strategy uses a straightforward method that first measures the performance of each individual demonstration (i.e., one-shot prompting) and then picks the top $K$ best demonstrations to create the final prompts. Note that Top-$K$ strategy assumes that the optimal prompt is typically built from demonstrations that perform well when used on their own. The Greedy search strategy follows the standard greedy search procedure, finding the best solution by making the best local choices at each step. At each step of the algorithm, the chosen demonstration is the one that allows the updated prompts to achieve the best performance.

To evaluate the effectiveness of our strategies, we conduct extensive experiments on various downstream tasks, such as foreground segmentation, single object detection, and colorization. Our results indicate that our method can significantly enhance the VFM's in-context learning performance in an effective and interpretable manner. The overall contribution is summarized as follows:

\begin{itemize}
\item We introduce a task-level prompt to avoid the significant computational cost and the risk of overfitting associated with sample-level methods, which select different demonstrations for different samples.
    
    \item We propose two time-saving and effective prompt search strategies to identify near-optimal prompts and achieve SOTA performance with minimal cost, which has not been achieved in prior work.
    
    \item The effectiveness of these two strategies is demonstrated in various tasks. While saving more than $98\%$ of the prompt searching time, consistent relative improvements of over $6.2\%$ are observed across different downstream tasks compared to state-of-the-art methods.
\end{itemize}

\section{Related works}

\subsection{Visual In-Context Learning}

The emergence of large language models (LLMs) like GPT-3~\cite{brown2020language}, BLOOM~\cite{workshop2022bloom}, and LLaMA~\cite{touvron2023llama} introduces a new learning paradigm, In-Context Learning(ICL)~\cite{8403294,9633155,wang2024large,baldassini2024makes}, which refers to the process of conditioning an LLM to solve various downstream NLP tasks using prompts constructed from a few demonstration input-output pairs~\cite{liu2022few,cho2023prompt,ma2024fairness} (i.e., few-shot prompting). In visual domain, MAE-VQGAN~\cite{bar2022visual} utilizes the model's grid inpainting capability to propose the first visual ICL model. Similarly, Painter~\cite{wang2023images} performs standard masked image modeling on the stitch of input and output image pairs to train an ICL model. Additionally, SegGPT~\cite{wang2023seggpt}, a variant of Painter, enhances the segmentation ability of the ICL model by randomly coloring similar semantic categories or objects. Previous works demonstrate the effectiveness of visual ICL models, and that selecting appropriate prompts ensures the model's understanding of task knowledge.

\subsection{Visual Prompt Selection}

To fully leverage the powerful reasoning capabilities of visual context learning models like MAE-VQGAN~\cite{bar2022visual} and Painter~\cite{wang2023images}, researchers are dedicated to exploring new algorithms for selecting appropriate prompts for different query samples(tasks)~\cite{huang2023diversity,ma2024learning}. These visual prompt selection methods can be divided into two categories, which are reward-model-based strategy and the rule-guided strategy: 
\begin{itemize}
    \item \textbf{Rule-guided} strategy methods include UnsupPR~\cite{NEURIPS2023_398ae57e} and prompt-SelF~\cite{sun2023exploring}. Both algorithms calculate the similarity between the query samples and the training samples using features extracted by the pretrained CLIP~\cite{radford2021learning} model to select demonstrations. The latter enhances the selected prompts to make full use of prompt information. However, such unsupervised algorithms do not utilize the labeled information of the training set and do not introduce task content constraints in the selection process, thus limiting the performance of the algorithms.
    \item \textbf{Reward-model-based} strategy methods include SupPR~\cite{NEURIPS2023_398ae57e} and InMeMo~\cite{zhang2024instruct}. Both algorithms train an additional scoring model on the training set to predict the compatibility between prompts and query samples. The latter proposes a prompt enhancer to improve the prompts and obtain a higher-performing scoring model. However, such supervised algorithms are costly to train, require a large amount of labeled data, and are prone to overfitting when the training samples are too few.
\end{itemize}

Therefore, to fully utilize the labeled information under few-shot supervision data, this paper proposes two simple greedy task-level prompt selection methods, which are Top-$K$ prompt selection method and Greedy prompt selection method. The former has a time complexity of $ O(N) $ ,and the latter has a time complexity of $ O(N^2) $ in the worst case. 

\section{Methods}

\subsection{Problem Setup}
Let $\mathcal{S}=\{(x_i,y_i)\}^N_{i=1}$ be a validation set consisting of $N$ image-label pairs, where $x_i$ denotes an image, and $ y_i $ is the corresponding label (e.g., $0/1$ masks in a segmentation task). The target of VICL is to select a subset of samples from $\mathcal{S}$ to create a demonstration set, denoted as $ \mathcal{P}=\{(x_i,y_i)\}^K_{i=1}$, which is used to prompt a pretrained foundation model $f$. This prompt aims to achieve the following goal: given a new query sample $(x_q,y_q)$, the foundation model should generate a prediction $\hat{y}_q=f(\mathcal{P},x_q)$ for $x_q$ that is as close as possible to the ground-truth $y_q$, which can be formally represented as minimizing $\mathcal{L}(\hat{y}_q, y_q)$, where $\mathcal{L}(\cdot)$ denotes the loss metric used for various specific tasks.

\subsection{Sample-level Prompt}

Sample-level prompt selection methods aim to find a $\mathcal{P}^*$ for each query sample:
\begin{equation}
  \mathcal{P}^*=\mathop{\mathrm{argmin}}\limits_{\mathcal{P} \subseteq \mathcal{S}}{\mathcal{L}(f(\mathcal{P},x_q),y_q)},
\end{equation}
Since $y_q$ is unknown, previous works~\cite{NEURIPS2023_398ae57e,zhang2024instruct,sun2023exploring} focus on constructing a scoring function $ g $ (which can be a manually set rule, like IOU, accuracy and so on, or parameters obtained from supervised learning) to automatically select the most suitable example(s) from the validation dataset $ S $ for a query sample $ x_q $. The prompt selection strategy is:
\begin{equation}
  x^*=\mathop{\mathrm{argmax}}\limits_{x\in \mathcal{S}}{g(x,x_q)}.
\end{equation}
These methods rank the training examples based on their scores and choose the top-$K$ example pairs. When $ K=1 $, they choose the optimal example pair as the prompt, $ \mathcal{P}_q=\{(x^*,y^*)\} $. 

Sample-level requires a significant amount of time to obtain a reward model (scoring model) and evaluate each query sample before VICL, which hinders the flexible deployment of VICL. However, these efforts seem to yield little benefit. In previous works, researchers find that differences in prompts can lead to significant variations in final performance, and intuitively search suitable prompts for different query samples. However, we discover a counter-intuitive phenomenon: although different prompts can cause dramatic changes in VICL's performance (as shown in Fig.~\ref{fig:motivation}(b)), the prompts that perform well on every different query sample are always the same (as shown in Fig.~\ref{fig:prompt-task}). Therefore, we only need to spend a small amount of computational cost to find the best task-level prompt, rather than spending a large amount of computational cost to determine the same prompt for all samples individually.


\subsection{Task-level Prompt}

As the prompts that perform well on every different query sample are always the same, we propose the objective of task-level prompt selection methods to construct a demonstration set $ \mathcal{P} $ from labeled data $ \mathcal{S} $ shared across different query samples $ x_q $: 
\begin{equation}
    \mathcal{P}^*=\mathop{\mathrm{argmin}}\limits_{\mathcal{P}\subseteq \mathcal{S}}{\mathop{\sum}\limits_{\mathcal{D}}{\mathcal{L}(f(\mathcal{P},x_q),y_q)}}, 
\end{equation}
where $\mathcal{D}$ includes all the unseen query samples, namely $ (x_q,y_q) \in \mathcal{D} $. Since obtaining $\mathcal{D}$ is difficult, we substitute $\mathcal{D}$ with known data $\mathcal{S}$ based on the generalization of dataset:
\begin{equation}
\mathcal{P}^*=\mathop{\mathrm{argmin}}\limits_{\mathcal{P}\subseteq \mathcal{S}}{\mathop{\sum}\limits_{\mathcal{S-P}}{\mathcal{L}(f(\mathcal{P},x_q),y_q)}}, 
\end{equation}
where $ \mathcal{S} \ne \mathcal{P} $, and $(x_q,y_q) \in \mathcal{S-P} $. The most straightforward approach in this method is to obtain all possible combinations of $ \mathcal{P} $ from $ \mathcal{S} $ and use the performance of $ \mathcal{P} $ on $ \mathcal{S}-\mathcal{P} $ to obtain loss. 

However, this approach still has a time complexity that is excessively high, with worst-case and best-case scenarios both being $ O(2^N) $ (we integrate the features of prompt set by summation pooling without considering the effect of order, and it will be $ O(N!) $ when considering order), making it difficult to achieve the optimal solution. Inspired by the method proposed in ~\cite{ma2024fairness}, which aims to address the prompt selection in LLM, this paper extends two simple task-level prompt selection methods, which are Top-$K$ prompt selection method and Greedy prompt selection method. The former has a time complexity of $ O(N) $ ,and the latter has a time complexity of $ O(N^2) $ in the worst case.

\subsubsection{Top-$K$ Prompt Selection Method}
The main idea of our Top-$K$ Prompt Selection Method proposed in this section is to simplify the basic idea of task-level prompt selection by using prior human knowledge to constrain the demonstration sets obtained from $\mathcal{S}$, appropriately reducing the combination possibilities to achieve a reduction in time complexity. Specifically, this means moving from using $O(2^N)$ time to obtain the optimal demonstration set to using $O(N)$ time to obtain the optimal single prompt:
\begin{equation}
x^*=\mathop{\mathrm{argmin}}\limits_{x \in \mathcal{S}}{\mathop{\sum}\limits_{\mathcal{S}-\{x\}}{\mathcal{L}(f(\{x\},x_q),y_q)}}, 
\end{equation}
where $(x_q,y_q) \in \mathcal{S-} \{x\}$. However, for a given task, a single prompt sometimes cannot provide enough information to complete the task, and multiple prompts are needed. Therefore, to obtain a demonstration set composed of multiple prompts, we sort the scores of the single prompts and select the top $K$ prompts to form the demonstration set. The specific algorithm is shown in Algorithm~\ref{alg:topk}.

Nevertheless, the Top-$K$ prompt selection method heavily relies on the choice of the hyperparameter $K$. When $K$ is too small, the information may be insufficient; when $K$ is too large, it may have a negative effect (as shown in Fig.~\ref{fig:two}). 
To automate the determination of the length of the demonstration set, this paper proposes a greedy prompt selection method.

\begin{minipage}{0.4\textwidth}
\begin{algorithm}[H]
\caption{Top-$K$ Prompt Selection Method}
\label{alg:topk}
\begin{algorithmic}[1] 
    \STATE \textbf{Given:} validation set $\mathcal{S}=\{(x_i,y_i)\}^N_{i=1} $, pretrained model $g$, number $K$
    \STATE Initial demonstration set $\mathcal{P}=\{\}$
    \FOR{$k$ in $1,...,K$}
    \STATE $x^*=\mathop{\mathrm{argmin}}\limits_{x \in \mathcal{S}}{\mathop{\sum}\limits_{\mathcal{S}-\{x\}}{\mathcal{L}(f(\{x\},x_q),y_q)}}$
    \STATE \emph{Insert} the top sample $x^*$ into $\mathcal{P}$ and \emph{remove} it from $\mathcal{S}$
    \ENDFOR
    \RETURN $\mathcal{P}$

\end{algorithmic}
\end{algorithm}
\end{minipage}

\subsubsection{Greedy Prompt Selection Method}
Our Greedy Prompt Selection Method proposed in this section follows the basic greedy approach, where at each stage, the current optimal prompt set is identified to guide the next stage's operation. This involves selecting the most promising new sample to add to the existing demonstration set based on:
\begin{equation}
\begin{aligned}
x^{greedy}& = \\
&\mathop{\mathrm{argmin}}\limits_{x\in(\mathcal{S}-\mathcal{P})}{\mathop{\sum}\limits_{\mathcal{S}-\mathcal{P}-\{x\}}{\mathcal{L}(f(\mathcal{P}+\{x\},x_q),y_q)}},
\end{aligned}
\label{eq:greedy}
\end{equation}
where $(x_q,y_q) \in \mathcal{S-P} -\{x\}$. Although the new sample represents the optimal solution for the current demonstration set, 
simply increasing the quantity of prompts does not necessarily improve performance and can even have adverse effects (as shown in Fig.~\ref{fig:two}). Therefore, it's crucial to compare the score $ a_{new}=g(\mathcal{P}+\{x\},\mathcal{S}-\mathcal{P}-\{x\}) $ obtained from adding the new sample with the score $ a_{ori}=g(\mathcal{P},\mathcal{S}-\mathcal{P}) $ of the original demonstration set. This comparison facilitates early stopping and pruning of the algorithm, aiming to achieve global optimal solution through local optimal solutions. When $ a_{ori} \le a_{new} $, it indicates that the new demonstration set performs better on the validation set, allowing the new sample to be included in demonstration set for the next iteration:
\begin{equation}
  \mathcal{P}=\mathcal{P}+\{x^{greedy}\}.
\end{equation}
However, when $ a_{ori} > a_{new} $, it suggests that adding the locally optimal sample no longer enhances the performance on the validation set, prompting the algorithm to halt the iteration and output the current demonstration set as the solution, $\mathcal{P}^*=\mathcal{P}$. The specific algorithm is illustrated in Algorithm~\ref{alg:greedy}.

\begin{minipage}{0.4\textwidth}
\begin{algorithm}[H]
\caption{Greedy Prompt Selection Method}
\label{alg:greedy}
\begin{algorithmic}[1] 
    \STATE \textbf{Given:} validation set $\mathcal{S}=\{(x_i,y_i)\}^N_{i=1} $, pretrained model $g$
    \STATE Initial demonstration set $\mathcal{P}=\{\}$
    \WHILE{$\mathcal{S}$ \emph{is not null}}
    \STATE Calculate the $x^{greedy}$ according to Eq.~\ref{eq:greedy}
    \STATE \emph{Insert} the greedy sample into $\mathcal{P}$ that can improve score best and \emph{remove} it from $\mathcal{S}$ 
     \STATE \emph{Stop} searching when score can't be improved
    \ENDWHILE
    \RETURN $\mathcal{P}$

\end{algorithmic}
\end{algorithm}
\end{minipage}

\begin{table*}[ht]
\centering
\caption{Main results. Our methods achieve the best performance and are close to the Oracle.}
\label{tab:main}
\resizebox{0.95\textwidth}{!}{
\begin{tabular}{c|c|c|c|ccccc} \toprule
\multirow{2}{*}{} 
& \multirow{2}{*}{\textbf{Method}}  
& \multirow{2}{*}{\textbf{Det.(mIOU)}$\uparrow$} 
& \multirow{2}{*}{\textbf{Color.(mse)}$\downarrow$}  
& \multicolumn{5}{c}{\textbf{Seg.(mIOU)}$\uparrow$} \\
& & & &\textbf{Split-0} &\textbf{Split-1} &\textbf{Split-2} &\textbf{Split-3} &\textbf{Avg.} 
\\ \midrule 

 \multirow{4}{*}{Sample} &{UnsupPR} &${24.01}_{0.53}$ &${64.95}_{0.27}$ &$\best{{37.28}_{1.19}}$ & ${39.44}_{1.87}$ &${33.32}_{3.04}$ &${28.22}_{0.59}$ &$34.57$ \\
 &{SupPR} &- &- &$\second{{36.97}_{1.25}}$ & ${39.74}_{1.78}$ &${34.24}_{1.88}$ &${29.15}_{1.05}$ &$35.03$ \\
    &{prompt-SelF} &${15.00}_{0.57}$ &$\best{{45.61}_{0.36}}$ &${32.76}_{1.18}$ &${38.45}_{0.92}$ &${35.77}_{0.23}$ &${29.38}_{0.20}$ &$34.09$\\
    &{SupPR-SelF} &- &- &${{33.42}_{0.96}}$ & ${38.97}_{0.44}$ &${35.64}_{0.46}$ &$\second{{30.40}_{0.37}}$ &$34.61$ \\
    \midrule 
    
 \multirow{5}{*}{Task}
     &{Oracle*} &$\textcolor{gray}{{29.03}_{2.84}}$ &$\textcolor{gray}{{61.56}_{1.27}}$ &$\textcolor{gray}{{39.09}_{0.77}}$ &$\textcolor{gray}{{44.37}_{0.98}}$ &$\textcolor{gray}{{37.93}_{0.42}}$ &$\textcolor{gray}{{32.40}_{1.06}}$ &$\textcolor{gray}{38.45}$\\
     \cmidrule{2-9} 
     &{Random} &${{24.53}_{0.88}}$ &${66.39}_{0.17}$ &${35.51}_{0.53}$ &${{39.83}_{2.39}}$ &${33.17}_{2.57}$ &${25.30}_{1.55}$ &$33.45$\\
     &{\text{Top-1}} &$\best{{28.25}_{2.94}}$ &$\second{{61.56}_{1.27}}$ &${32.41}_{2.31}$ &$\best{{42.22}_{1.42}}$ &$\second{{37.00}_{0.82}}$ &${30.20}_{2.44}$ &$\second{35.46}$\\
    &{Greedy} &$\best{{28.25}_{2.94}}$ &$\second{{61.56}_{1.27}}$ &${36.86}_{1.38}$ &$\best{{42.22}_{1.42}}$ &$\best{{37.11}_{0.87}}$ &$\best{{30.84}_{1.84}}$ &$\best{36.76}$\\

 \bottomrule
\end{tabular}
}
\end{table*}

\section{Experiments}
In this section, we discuss the performance comparison between the two proposed task-level prompt selection methods and other unsupervised prompt selection methods, as well as the performance impact of different methods for constructing $\mathcal{S}$. Additionally, we explore the necessity of using a greedy strategy to achieve a global solution, and conduct fine-grained, in-depth experiments starting from individual samples.


\begin{figure}[ht]
     \begin{subfigure}[b]{0.45\linewidth}
         \centering
         \includegraphics[width=\textwidth]{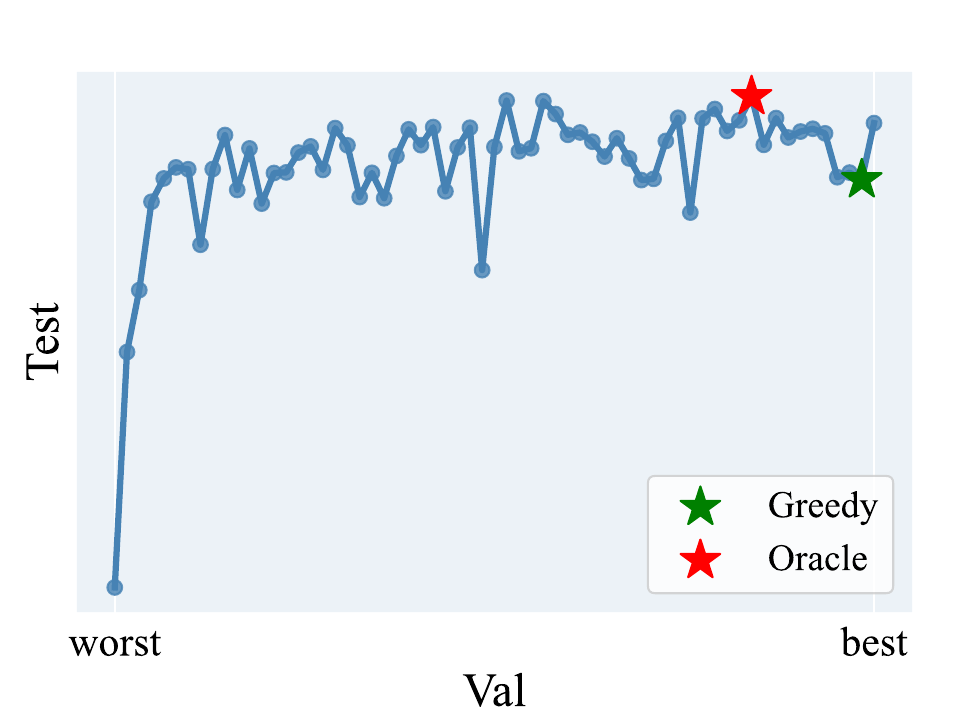}
         \caption{split-0}
     \end{subfigure}
     \hfill
     \begin{subfigure}{0.45\linewidth}
     \includegraphics[width=\textwidth]{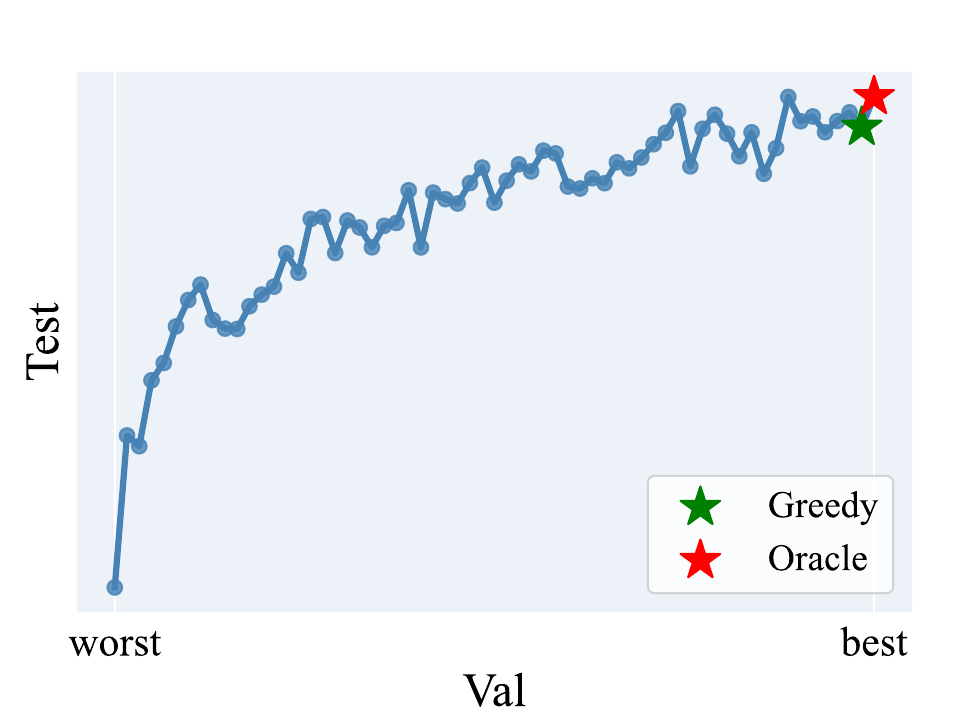}
     \caption{split-1}
     \end{subfigure} \\
     \begin{subfigure}[b]{0.45\linewidth}
         \centering
          \includegraphics[width=\textwidth]{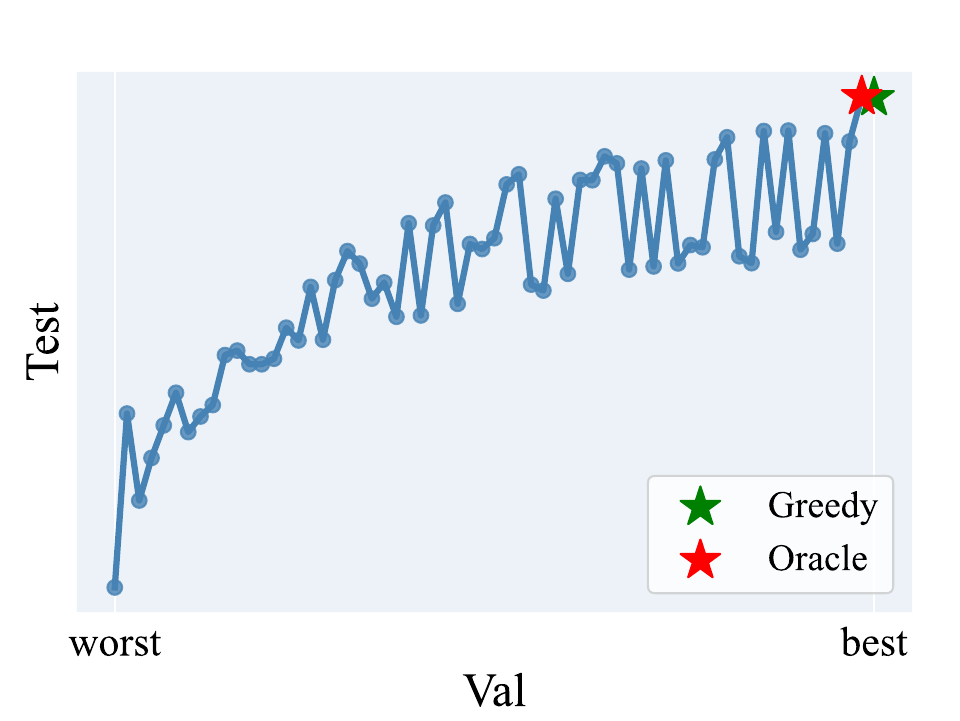}
         \caption{split-2}
     \end{subfigure}
     \hfill
     \begin{subfigure}{0.45\linewidth}
     \includegraphics[width=\textwidth]{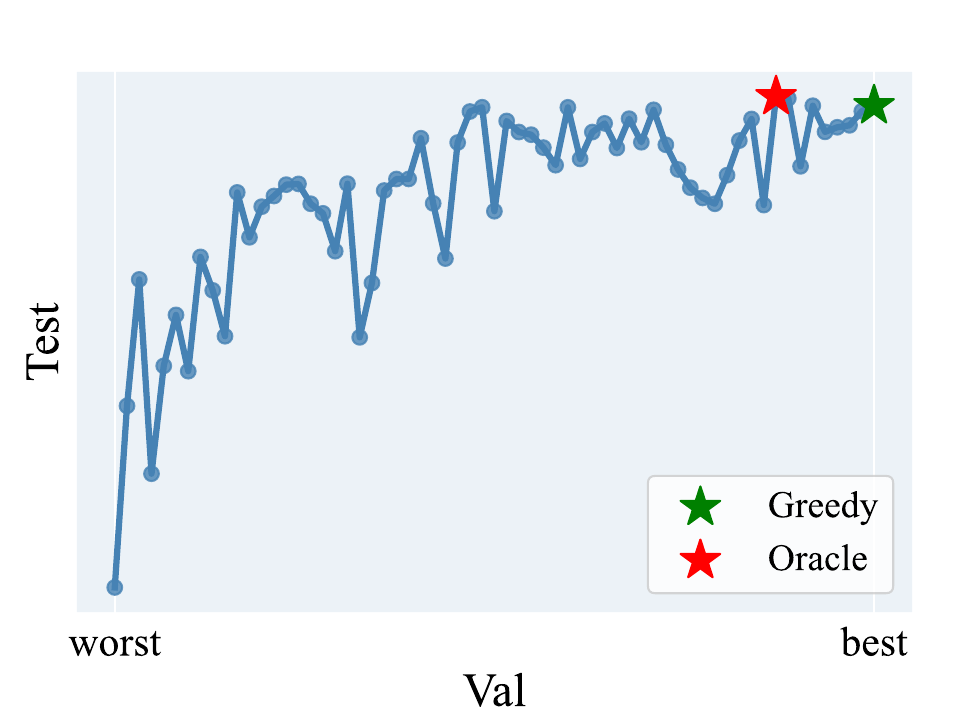}
     \caption{split-3}
     \end{subfigure}
     \caption{Performance of all prompts on test set, where ``Greedy'' and ``Oracle'' indicate performance under prompts selected by our strategy and the upper-bound performance across all prompts, respectively.}
     \label{fig:seg}
 \end{figure}

\subsection{Setup}



According to~\cite{bar2022visual,NEURIPS2023_398ae57e,sun2023exploring}, we conduct few-shot evaluations on three out-of-distribution (OOD) computer vision tasks:
\begin{itemize}
    \item \textbf{Foreground Segmentation.} We use the Pascal-5i~\cite{shaban2017one} dataset, which is comprised of 4 different image splits, where each split contains data from 5 categories. For evaluation, we report the mean Intersection Over Union (mIOU) metric.

    \item \textbf{Single Object Detection.} We use the Pascal VOC 2012~\cite{everingham2015pascal} dataset, which contains images and their associated detection boxes. The evaluation method is similar to Foreground Segmentation, and we report the mIOU metric.

    \item \textbf{Colorization.} We use a subset of the ImageNet~\cite{russakovsky2015imagenet} dataset, which contains data from 1000 categories. We randomly sample 50,000 example pairs and image queries from the ImageNet validation set. For evaluation, we report the Mean Squared Error (MSE) loss (scaled up by a factor of 100).
\end{itemize}

In each task, the few-shot in-context examples come from the training set, with a default size of $N=16$. For all experiments, we perform evaluations on a pre-trained image inpainting model, MAE-VQGAN~\cite{bar2022visual}, which consists of an encoder and a decoder. When the prompt has $K$ samples, each sample is combined with the query image into a grid of 2×2 sub-images. These are then passed through the encoder to obtain $K$ features, which are summed to get fused features and then input to the decoder for the final output.


\begin{figure}[ht]
    \centering
     \begin{subfigure}[b]{0.75\linewidth}
         \centering
         \includegraphics[width=\textwidth]{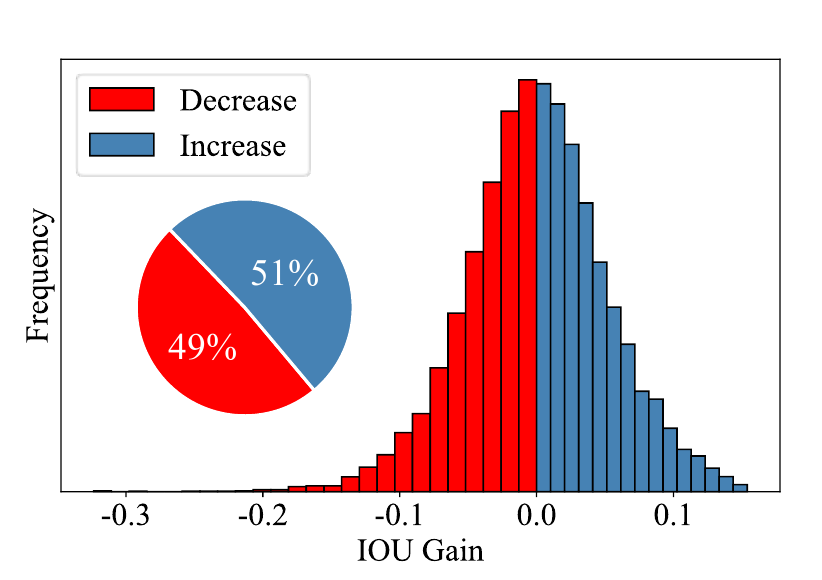}
     \end{subfigure}
     \caption{Performance of the demonstration set with a length of 2 and its subsets for each sample, where half of the combinations shows a performance drop when adding samples.}
     \label{fig:two}
 \end{figure}


\begin{figure*}[ht]
    \centering
     \begin{subfigure}{0.90\textwidth}
     \includegraphics[width=\textwidth]{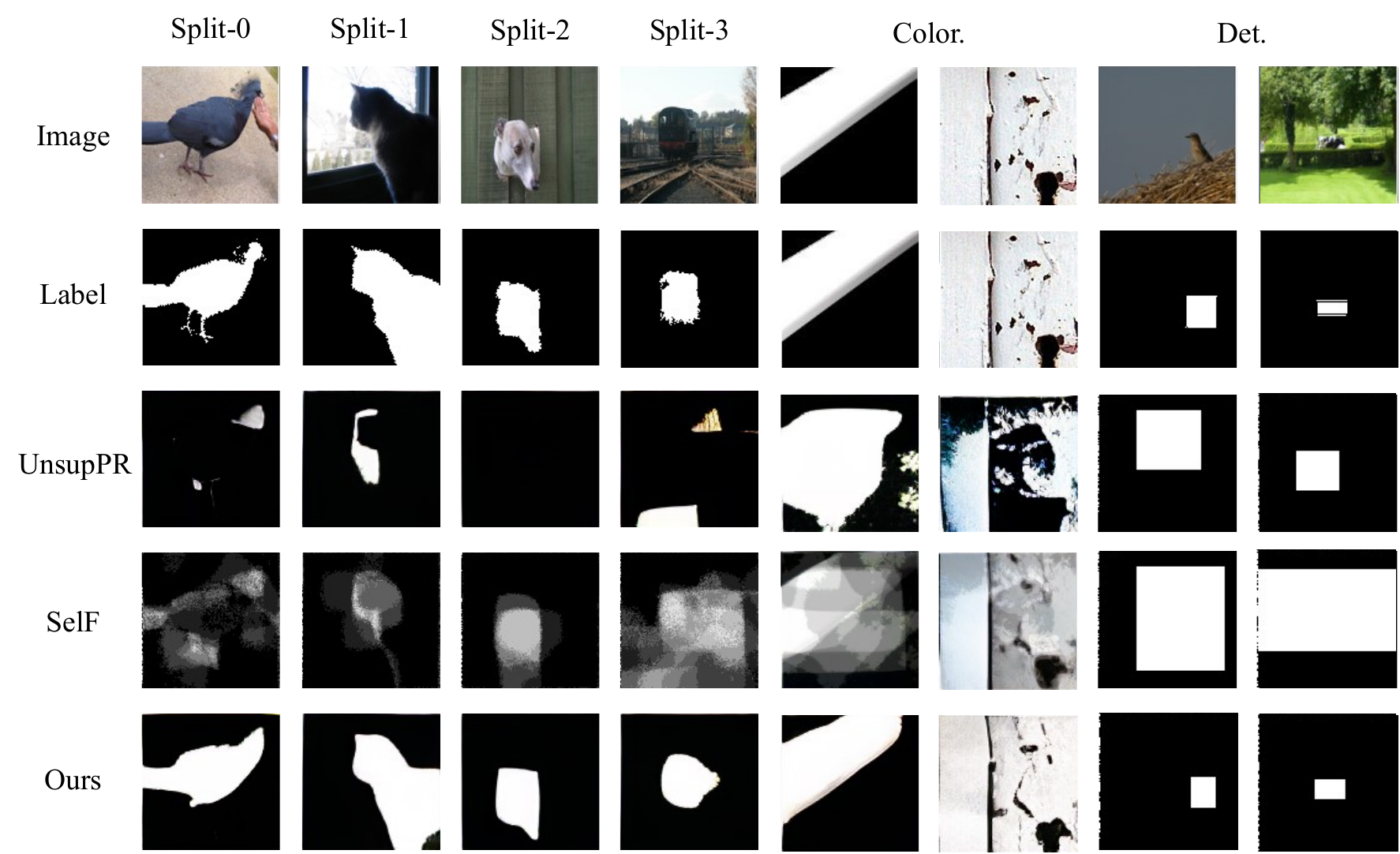}
     \end{subfigure}
     \caption{ In-context results retrieved by several unsupervised prompt selection methods, where ``Ours'' is the Greedy Prompt Selection Method. }
     \label{fig:sample}
 \end{figure*}


\begin{figure}[ht]
    \centering
     \begin{subfigure}{0.85\linewidth}
     \includegraphics[width=\textwidth]{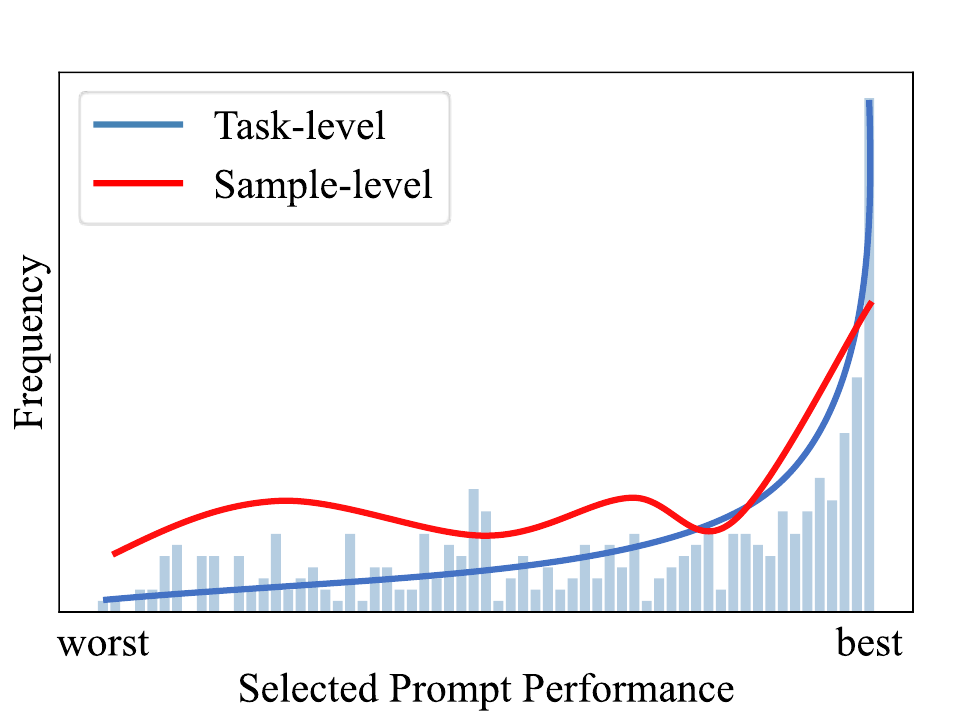}
     \end{subfigure}
     \caption{A comparison of the performance of prompts selected by different strategies among all possible prompts (the far right indicates that the selected prompt is the best-performing one among all possible prompts).}
     \label{fig:prompt-task}
 \end{figure}

In the experiments, we mainly compare six unsupervised prompt selection methods:
\begin{itemize}
    \item \textbf{Random.} Randomly select prompt combinations from the few-shot in-context examples. (The result of the random method is obtained by averaging the performance of all possible few-shot in-context examples combinations on the test set.)\footnote{$O(2^N)$ when $N=16$, we randomly select $N'=6$ samples for obtaining $\mathcal{P}$ and use the remaining $N-N'=10$ samples for validation set performance evaluation in the random, Oracle, and our methods. }

    \item \textbf{Oracle.} The task-level best-performing combination on the test set from all possible few-shot in-context examples combinations (the upper bound performance of task-level prompting).

    \item \textbf{Top-$K$.} Select the top-$K$-performing samples under one-shot VICL on the validation set.
    
    \item \textbf{Greedy.} Use a greedy method to select from the few-shot in-context examples.

    \item \textbf{UnsupPR.}~\cite{NEURIPS2023_398ae57e} Unsupervised prompt retrieval (UnsupPR), a method that uses off-the-shelf features for nearest example search. We use the cosine similarity between features extracted by CLIP’s vision encoder to search for the nearest example.

    \item \textbf{Prompt-SelF.}~\cite{sun2023exploring} Use the nearest example obtained from UnsupPR as the query image and fuses the query image and prompt image using different arrangements to create 8 new fused images.
\end{itemize}
Also, we compare two supervised prompt selection methods:
\begin{itemize}
    \item \textbf{SupPR.}~\cite{NEURIPS2023_398ae57e} Supervised Prompt Retrieval(SupPR), is a supervised prompt retrieval method, which trains a neural network to choose examples that directly maximize in-context learning performance.
    \item \textbf{SupPR-SelF.} It is a combined method of SupPR and Prompt-SelF, which uses SupPR to select prompt and Prompt-SelF to do augment of prompt.
\end{itemize}
The eight methods are mainly divided into two categories: UnsupPR, Prompt-SelF, SupPR and SupPR-SelF belong to sample-level prompt selection methods, while Random, Oracle, and Ours (including Top-$K$ and Greedy) belong to task-level prompt selection methods.

\subsection{Results}

\begin{table*}[ht]
\centering
\caption{Sort each sample based on its performance on the remaining validation set, and select the top-$K$ as the target prompt set, where $K=\{1,2,4\}$.}
\label{tab:ablation}
\begin{tabular}{c|c|c|ccccc} \toprule
\multirow{2}{*}{\textbf{Method}}  
& \multirow{2}{*}{\textbf{Det.(mIOU)}$\uparrow$} 
& \multirow{2}{*}{\textbf{Color.(mse)}$\downarrow$}  
& \multicolumn{5}{c}{\textbf{Seg.(mIOU)}$\uparrow$} \\
& & &\textbf{Split-0} &\textbf{Split-1} &\textbf{Split-2} &\textbf{Split-3} &\textbf{Avg.} 
\\ \midrule 
{Oracle*} &$\textcolor{gray}{{29.03}_{2.84}}$ &$\textcolor{gray}{{61.56}_{1.27}}$ &$\textcolor{gray}{{39.09}_{0.77}}$ &$\textcolor{gray}{{44.37}_{0.98}}$ &$\textcolor{gray}{{37.93}_{0.42}}$ &$\textcolor{gray}{{32.40}_{1.06}}$ &$\textcolor{gray}{38.45}$\\
\midrule
{Random} &${24.53}_{0.88}$ &${66.39}_{0.17}$ &${35.51}_{0.53}$ &${39.83}_{2.39}$ &${33.17}_{2.57}$ &${25.30}_{1.55}$ &$33.45$\\
{\text{Top-1}} &$\bestcls{{28.25}_{2.94}}$ &$\bestcls{{61.56}_{1.27}}$ &${32.41}_{2.31}$ &${42.22}_{1.42}$ &${37.00}_{0.82}$ &${30.20}_{2.44}$ &$35.46$\\
{\text{Top-2}} &${26.55}_{0.41}$ &${63.70}_{0.95}$ &${36.86}_{1.22}$ &${43.07}_{1.13}$ &$\bestcls{{37.56}_{0.29}}$ &${29.74}_{1.15}$ &$36.81$\\
{\text{Top-4}} &${26.72}_{2.22}$ &${65.66}_{0.50}$ &$\bestcls{{37.71}_{0.68}}$ &$\bestcls{{43.36}_{1.15}}$ &${36.43}_{1.17}$ &${30.24}_{1.12}$ &$\bestcls{36.94}$\\
    
    {Greedy} &$\bestcls{{28.25}_{2.94}}$ &$\bestcls{{61.56}_{1.27}}$ &${36.86}_{1.38}$ &${{42.22}_{1.42}}$ &${{37.11}_{0.87}}$ &$\bestcls{{30.84}_{1.84}}$ &${36.76}$\\

 \bottomrule
\end{tabular}
\end{table*}

\begin{table}[!ht]
\centering
\caption{A comparison of the worst-case complexity of different methods, where $N$ and $M$ represent the sizes of the demonstration set and test set, respectively ($N \ll M$).}
\label{tab:time}
\begin{tabular}{c|c|c} \toprule
\multirow{1}{*}{} 
& \multirow{1}{*}{\textbf{Method}}  
& \multirow{1}{*}{\textbf{Complexity}}\\ \midrule 

 \multirow{4}{*}{Sample} &{UnsupPR} &$O(NM)$ \\
 &{SupPR} &Train-required \\
    &{prompt-SelF} &$O(NM)$ \\
    &{SupPR-SelF} &Train-required \\
    \midrule 
    
 \multirow{2}{*}{Task}
     &{\text{Top-1}} &$O(N)$ \\
    &{Greedy} &$O(N^2)$ \\

 \bottomrule
\end{tabular}
\end{table}


    


We conduct experiments on different settings and report the results of three runs (with $seed=\{0, 1, 2\}$). In Tab.~\ref{tab:main}, it presents the main results of the experiments. The following findings can be observed: (1) Our methods achieve state-of-the-art performance across various datasets, both when compared to sample-level methods and task-level methods, demonstrating the effectiveness of our approaches. In Tab.~\ref{tab:main}, our methods achieve optimal results in detection and segmentation tasks, and even achieve a global optimal solution in the coloring task. (2) Under certain conditions, our methods can reach the global optimal solution, with overall results very close to Oracle. In the detection task, our Greedy method is less than $3\%$ away from Oracle, and in the segmentation task, the average results across different splits are less than $6\%$ away from Oracle. Additionally, in the coloring task, the results achieve the global optimal solution, consistent with Oracle (even if the results did not reach the optimal in the comparison of different methods). In Fig.~\ref{fig:seg}, it shows that the results of the validation set and the test set are highly consistent, and our Greedy method can obtain results very close to the global optimal solution. (3) As shown in Tab.~\ref{tab:time}, our task-level methods not only achieve state-of-the-art performance, but also have the lowest worst-case complexity.

\textbf{Task-level vs. Sample-level.} As shown in Fig.~\ref{fig:motivation}(d), the task-level optimal prompt enables $27\%$ of the samples to achieve the optimal solution, while current sample-level prompt selection methods can only find the optimal solution for $15.03\%$ of the samples. Moreover, we conduct a detailed study comparing the performance of prompts selected by the task-level method and the sample-level method on individual samples. As illustrated in Fig.~\ref{fig:prompt-task}, we compare the selected prompt with all possible demonstration sets on individual samples and plot the ranking frequency of these samples across the entire test dataset. 
It can be observed that both the task-level and sample-level methods exhibit long-tail distributions in their frequency curves, indicating that under a specific task, the prompts resulting in the best performance for different samples are always the same. However, the distribution of the task-level method is significantly more concentrated than that of the sample-level method, with frequency increasing more rapidly as performance improves. It demonstrates that the simple task-level method is more effective than the complex sample-level method.
The prompt that performs best on the task, even if it cannot achieve optimal performance for some samples, can still obtain relatively good results compared with most prompts. In other words, there is no need to spend a massive amount of computational effort at the risk of overfitting to select different demonstrations to form a prompt for distinct samples.

  \textbf{Top-$K$ vs. Greedy.} Based on the simple assumption that the optimal prompt is typically built from demonstrations that perform well, the most straightforward approach is to sort each sample based on its performance on the remaining validation set and select the top $K$ as the target demonstration set, which has a lower time complexity of $O(N)$. Therefore, in this section, we compare the Greedy method with the Top-$K$ method in Tab.~\ref{tab:ablation}. The Top-$K$ method is also effective, and when $K=1$, its results are very similar to Greedy (except for split-0 in the segmentation task). When comparing different $K$ values, we can observe that for the segmentation task, a larger $K$ value can lead to better results. However, simply choosing a higher $K$ value does not necessarily improve performance shown in Figure~\ref{fig:two}. 
  Particularly in the detection and colorization tasks, increasing the $K$ value results in a significant performance decline. This illustrates a limitation of the Top-$K$ method: it cannot reliably determine an appropriate hyperparameter $K$. Our proposed Greedy method, on the other hand, avoids this issue by adaptively determining the length of the demonstration set.

We visualize the in-context results of different datasets under various unsupervised prompt selection methods in Fig.~\ref{fig:sample}, which are most distinguishable. Additionally, the association between UnsupPR and SelF methods is noticeable. It is evident that some prompts selected by UnsupPR cause the model to find shortcuts, resulting in outputs that have low relevance to the query image but high relevance to the prompt image. For instance, in the coloring task, the outputs of UnsupPR are severely distorted. Consequently, since UnsupPR performs poorly on these samples, SelF, which is an enhanced method based on UnsupPR, also performs poorly, resulting in scattered and messy integrated results. These sample results partially reflect the robustness of our method, but also reveal its limitations. The selected prompts do not handle some details well, such as bird legs and wall cracks.

\section{Conclusion}

In this paper, based on the observation that most test samples achieve optimal performance under the same prompt, we propose the task-level prompt strategy that significantly reduces inference computational costs. Furthermore, we introduce two train-free demonstration search strategy which can identify a near-optimal combination of demonstrations with minimal computational cost. Comprehensive experiments validate the effectiveness of our proposed method, demonstrating its ability to identify better demonstration combinations at a reduced cost compared to previous methods. These insights hold great promise for the further development and application of VICL, paving the way for more efficient and cost-effective paradigms.

\section*{Acknowledgement}

This work is supported by the National Natural Science Foundation of China (Grant No.62376193)

\bibliography{aaai25}

\end{document}